\documentclass[10pt,twocolumn,letterpaper]{article}

\usepackage[pagenumbers]{cvpr} 
\usepackage[accsupp]{axessibility}

\usepackage[dvipsnames]{xcolor}

\usepackage{mathtools}
\usepackage{amsthm}
\usepackage{mathrsfs}
\usepackage{bbm}
\usepackage[ruled]{algorithm2e}
\usepackage[pagebackref,breaklinks,colorlinks,citecolor=cvprblue]{hyperref}
\usepackage{url}
\usepackage{pifont}
\usepackage{multirow}

\usepackage{nicefrac}
\usepackage[capitalise]{cleveref}
\usepackage{makecell}
\graphicspath{{Figures/}}

\theoremstyle{plain}
\newtheorem{theorem}{Theorem}[section]
\newtheorem{proposition}[theorem]{Proposition}
\newtheorem{lemma}[theorem]{Lemma}
\newtheorem{corollary}[theorem]{Corollary}
\theoremstyle{definition}
\newtheorem{definition}[theorem]{Definition}

\theoremstyle{remark}
\newtheorem{remark}[theorem]{Remark}

\providecommand{\calM}{\mathcal{M}}
\providecommand{\calN}{\mathcal{N}}

\providecommand{\bbD}{\mathbb{D}}
\providecommand{\sym}[1]{\mathcal{S}^{#1}}
\providecommand{\spd}[1]{\mathcal{S}^{#1}_{++}}

\providecommand{\tril}[1]{\mathcal{L}^{#1}}
\providecommand{\bbR}[1]{\mathbb {R}^{#1}}
\providecommand{\bbRplus}{\mathbb {R}_{+}}
\providecommand{\bbRscalar}{\mathbb {R}}
\providecommand{\cinf}{C^{\infty}}
\providecommand{\orth}[1]{\mathrm{O}({#1})}
\providecommand{\bfst}{\mathbf{ST}}

\providecommand{\rieexp}{\operatorname{Exp}}
\providecommand{\rielog}{\operatorname{Log}}
\providecommand{\diff}{\operatorname{d}}
\providecommand{\diffphi}[1]{\phi_{*,#1}}
\providecommand{\diffphiinv}[1]{\phiinv_{*,#1}}

\providecommand{\pt}[2]{\Gamma_{#1 \rightarrow #2}}

\providecommand{\bbD}{\mathbb {D}}

\providecommand{\mln}{\phi_{mln}}

\providecommand{\mlog}{\mathrm{mlog}}

\providecommand{\mexp}{\mathrm{mexp}}

\providecommand{\phiinv}{\phi^{-1}}

\providecommand{\gphi}{g^{\phi}}
\providecommand{\phiMul}{\odot_{\phi}}
\providecommand{\diff}{\operatorname{d}}
\providecommand{\tr}{\operatorname{tr}}

\providecommand{\geuc}{g^{\mathrm{E}}}

\providecommand{\gphi}{g^{\phi}}

\providecommand{\dphi}{d^{\phi}}

\providecommand{\tr}{\operatorname{tr}}
\providecommand{\chol}{\operatorname{Chol}}

\providecommand{\dlog}{\operatorname{Dlog}}

\providecommand{\bzero}{\mathbf{0}}

\providecommand{\rmF}{\mathrm{F}}

\providecommand{\lem}{\mathrm{LEM}}
\providecommand{\lcm}{\mathrm{LCM}}
\providecommand{\st}{\mathrm{s.t.}}
\providecommand{\alphabeta}{(\alpha,\beta)}

\providecommand{\biparamAIM}{(\alpha,\beta)\text{-AIM}}
\providecommand{\biparamLEM}{(\alpha,\beta)\text{-LEM}}
\providecommand{\triparamLEM}{(\theta,\alpha,\beta)\text{-LEM}}

\providecommand{\paramLCM}{(\theta)\text{-LCM}}

\providecommand{\pow}[1]{\operatorname{Pow}_{#1}}

\providecommand{\ie}{\emph{i.e.,}}

\crefname{equation}{Eq.}{Eqs.}
\Crefname{equation}{Equation}{Equations}
\crefname{figure}{Fig.}{Figs.}
\Crefname{figure}{Figure}{Figures}
\crefname{table}{Tab.}{Tabs.}
\Crefname{table}{Table}{Tables}
\crefname{section}{Sec.}{Secs.}
\Crefname{section}{Section}{Sections}
\crefname{appendix}{App.}{Apps.}
\Crefname{appendix}{Appendix}{Appendices}

\crefname{theorem}{Thm.}{Thms.}
\Crefname{theorem}{Theorem}{Theorems}
\crefname{lemma}{Lem.}{Lems.}
\Crefname{lemma}{Lemma}{Lemmas}
\crefname{definition}{Def.}{Defs.}
\Crefname{definition}{Definition}{Definitions}
\crefname{corollary}{Cor.}{Cors.}
\Crefname{corollary}{Corollary}{Corollaries}
\crefname{remark}{Rem.}{Rems.}
\Crefname{remark}{Remark}{Remarks}
\crefname{proposition}{Prop.}{Props.}
\Crefname{proposition}{Proposition}{Propositions}

\definecolor{cvprblue}{rgb}{0.21,0.49,0.74}

\def\paperID{5737} 
\def\confName{CVPR}
\def\confYear{2024}

\title{Riemannian Multinomial Logistics Regression for SPD Neural Networks}

\author{
Ziheng Chen$^1$, 
Yue Song$^1$\thanks{Corresponding author.},
Gaowen Liu$^2$, 
Ramana Rao Kompella$^2$,
Xiao-Jun Wu$^3$
\& Nicu Sebe$^1$\\
$^1$ University of Trento, $^2$ Cisco Systems, $^3$ Jiangnan University\\
\tt{ziheng\_ch@163.com, yue.song@unitn.it}
}

\begin{document}
\maketitle

\begin{abstract}
Deep neural networks for learning Symmetric Positive Definite (SPD) matrices are gaining increasing attention in machine learning.
Despite the significant progress, most existing SPD networks use traditional Euclidean classifiers on an approximated space rather than intrinsic classifiers that accurately capture the geometry of SPD manifolds. Inspired by Hyperbolic Neural Networks (HNNs), we propose Riemannian Multinomial Logistics Regression (RMLR) for the classification layers in SPD networks. 
We introduce a unified framework for building Riemannian classifiers under the metrics pulled back from the Euclidean space, and showcase our framework under the parameterized Log-Euclidean Metric (LEM) and Log-Cholesky Metric (LCM).
Besides, our framework offers a novel intrinsic explanation for the most popular LogEig classifier in existing SPD networks. 
The effectiveness of our method is demonstrated in three applications: radar recognition, human action recognition, and electroencephalography (EEG) classification.    
The code is available at \url{https://github.com/GitZH-Chen/SPDMLR.git}.
\end{abstract}

\section{Introduction}
\label{sec:intro}
Symmetric Positive Definite (SPD) matrices are commonly encountered in a diverse range of scientific fields, such as medical imaging \citep{chakraborty2018statistical,chakraborty2020manifoldnet}, signal processing \citep{arnaudon2013riemannian, hua2017matrix,brooks2019exploring, brooks2019riemannian}, elasticity \citep{moakher2006averaging, guilleminot2012generalized}, question answering \citep{lopez2021vector,nguyen2022gyro}, graph classification \cite{chen2023distribution}, and computer vision \citep{huang2017riemannian,harandi2018dimensionality,zhen2019dilated, chakraborty2020manifoldnorm,chen2020covariance,zhang2020deep,chakraborty2020manifoldnet,chen2021hybrid,song2021approximate,nguyen2021geomnet,nguyen2022gyrovector,song2022fast}.
Despite their ubiquitous presence, traditional learning algorithms are ineffective in handling the non-Euclidean geometry of SPD matrices.
To address this limitation, several Riemannian metrics have been proposed, including Affine-Invariant Metric (AIM) \citep{pennec2006riemannian}, Log-Euclidean Metric (LEM) \citep{arsigny2005fast}, and Log-Cholesky Metric (LCM) \citep{lin2019riemannian}.
With these Riemannian metrics, various machine learning techniques can be generalized into SPD manifolds.

Inspired by the great success of deep learning \citep{hochreiter1997long, krizhevsky2012imagenet,he2016deep}, several deep networks have been developed on SPD manifolds.
Despite their promising performance, many approaches still rely on Euclidean spaces for classification, such as tangent spaces \citep{huang2017riemannian, brooks2019riemannian,nguyen2021geomnet,wang2021symnet,nguyen2022gyro,nguyen2022gyrovector,kobler2022spd,wang2022dreamnet,chen2023riemannian}, ambient Euclidean spaces \citep{wang2020deep, song2021approximate, song2022eigenvalues}, and coordinate systems \citep{chakraborty2018statistical}.
However, these strategies distort the intrinsic geometry of the SPD manifold, undermining the effectiveness of SPD neural networks. 
Notably, there are also some similarity-based classifiers originally designed for shallow learning methods \cite{gao2019robust,harandi2018dimensionality,chen2021hybrid}.
Although these classifiers can be extended to deep SPD neural networks \cite{wang2022learning,wangspdmetric2024}, the calculation of pair-wise distance might undermine the training efficiency.
Recently, motivated by HNNs \citep{ganea2018hyperbolic}, three kinds of SPD Multinomial Logistics Regression (MLR) based on the gyro-structures induced by LEM, LCM and AIM are developed in \cite{nguyen2023building}.
However, the proposed SPD MLRs rely on the gyro-structures, limiting their generality.
Besides, in \cite{chakraborty2020manifoldnet}, the authors also introduce an invariant layer for manifold-valued data mimicking the invariant FC layer in CNNs. However, it is designed for gridded manifold-valued data, which is not the primary data type encountered in many other SPD networks. Following the convention of most SPD networks, we only focus on non-gridded cases.

\textit{In fact, SPD MLR can be directly derived under LEM and LCM without the assistance of gyro structures.} More generally, LEM and LCM belong to Pullback Euclidean Metrics (PEMs), which are metrics pulled back from the Euclidean space.
This paper focuses on PEMs and proposes a unified framework for building SPD Multinomial Logistics Regression (SPD MLR) under PEMs. 
On the empirical side, we focus on the parameterized Log-Euclidean Metric (LEM) and Log-Cholesky Metric (LCM) \cite{chen2024liebn}, which generalize the standard LEM and LCM by the pullback of matrix power.
We showcase our SPD MLRs under these parameterized metrics.
Besides, our framework encompasses the gyro SPD MLRs induced by the standard LEM and LCM in \cite{nguyen2023building}.
More importantly, our framework also provides an intrinsic explanation for the commonly used LogEig classifier on SPD manifolds, which consists of successive matrix logarithm, FC, and softmax layers. 
Finally, extensive experiments demonstrate that our proposed Riemannian classifiers exhibit consistent performance gains across widely used SPD benchmarks.
The main \textbf{contributions} are summarized as follows: 
\begin{enumerate}[label={(\alph*)},itemsep=2pt,topsep=0pt,parsep=0pt]
    \item 
    We introduce a general framework for building SPD MLRs under PEMs and design specific SPD MLRs under two parameterized metric families. 
    \item
    Our framework offers an intrinsic explanation of the most popular LogEig classifier which stacks matrix logarithm, the FC layer, and softmax.
    \item 
    Extensive experiments on widely used SPD learning benchmarks demonstrate the superiority of our proposed classifiers over the previous baselines.
\end{enumerate}

\noindent\textbf{Main theoretical results: }
\cref{def:spd_hyperplane,def:spd_mlr} introduce the definitions of the SPD hyperplane and SPD MLR, respectively. The core idea lies in the computation of marginal distance to the hyperplane defined in \cref{eq:spd_pem_dist_hyperplane}. As \cref{lem:dist_to_hyperplane_pems} demonstrates, this problem admits a closed-form solution under any PEM.
Consequently, we establish a uniform expression of SPD MLR under any PEM in \cref{thm:general_mlr_pems}.
As the parameterized LEM and LCM all belong to PEMs, the associated SPD MLRs can be obtained by \cref{thm:general_mlr_pems}, the expressions of which are presented in \cref{cor:spd_mlr_param_lem_lcm}.
Finally, our framework also offers an intrinsic explanation for the widely used LogEig classifier in \cref{prop:equivalence_lem_mlr}. 
Due to page limits, all the proofs are placed in \cref{app:sec:proof}.
\section{Preliminaries}
\label{sec:geom_spd}
This section briefly reviews some basic concepts in Riemannian geometry and SPD manifolds. 
Please refer to \cite{do1992riemannian,loring2011introduction} for in-depth discussions.

\subsection{Riemannian geometry}
We first recap the concept of the pullback metric, which is ubiquitous in differential manifolds.
\begin{definition} [Pullback Metrics] \label{def:pullback_metrics}
    Suppose $\calM,\calN$ are smooth manifolds, $g$ is a Riemannian metric on $\calN$, and $f:\calM \rightarrow \calN$ is a diffeomorphism.
    Then $f$ can induce a Riemannian metric on $\calM$ defined as
    \begin{equation}
        (f^*g)_p(V_1,V_2) = g_{f(p)}(f_{*,p}(V_1),f_{*,p}(V_2)),
    \end{equation}
    where $p \in \calM$, $f_{*,p}(\cdot)$ is the differential map of $f$ at $p$, $V_i \in T_p\calM$, and $f^*g$ is the pullback metric by $f$ from $\calN$.
\end{definition}

The exponential \& logarithmic maps and parallel transportation are also crucial for Riemannian approaches in machine learning.
To bypass the notation burdens caused by their definitions, we review the geometric reinterpretation of these operators \cite{pennec2006riemannian, do1992riemannian}.
In detail, in a manifold $\calM$, geodesics correspond to straight lines in Euclidean space.
A tangent vector $\overrightarrow{x y} \in T_x\calM$ can be locally identified to a point $y$ on the manifold by geodesic starting at $x$ with an initial velocity of $\overrightarrow{x y}$, \ie $y=\rieexp_x(\overrightarrow{x y})$.
On the other hand, the logarithmic map is the inverse of the exponential map, generating the initial velocity of the geodesic connecting $x$ and $y$, \ie $\overrightarrow{x y}=\rielog_x(y)$.
These two operators generalize the idea of addition and subtraction in the Euclidean space.
For the parallel transportation $\pt{x}{y}(V)$, it is a generalization of parallelly moving a vector along a curve in Euclidean space.
we summarize the reinterpretation in \cref{tb:reinter_riem_operators}.

\begin{table}[htbp]
    \centering
    \resizebox{\linewidth}{!}{
    \begin{tabular}{ccc}
    \toprule
    Operations & Euclidean spaces & Riemannian manifolds \\
    \midrule
    Straight line & Straight line & Geodesic \\
    Subtraction & $\overrightarrow{x y}=y-x$ & $\overrightarrow{x y}=\log _x(y)$ \\
    Addition & $y=x+\overrightarrow{x y}$ & $y=\exp _x(\overrightarrow{x y})$ \\
    Parallelly moving & $V \rightarrow V$ & $\pt{x}{y}(V)$\\
    \bottomrule
    \end{tabular}
    }
    \caption{Reinterpretation of Riemannian operators.}
    \label{tb:reinter_riem_operators}
    \vspace{-3mm}
\end{table}

\subsection{The geometry of SPD manifolds}
Now, we introduce some necessary preliminaries about SPD manifolds.
The set of SPD matrices, denoted as $\spd{n}$, forms a smooth manifold known as the SPD manifold \citep{arsigny2005fast}. 
Several successful Riemannian metrics have been established on SPD manifolds, such as LEM \citep{arsigny2005fast}, AIM \citep{pennec2006riemannian} and LCM \citep{lin2019riemannian}. 
Recently, LEM and AIM are generalized into two-parameter families of metrics \cite{thanwerdas2023n} , namely $\biparamAIM$ and $\biparamLEM$ by the $\orth{n}$-invariant inner product on the Euclidean space $\sym{n}$ of symmetric matrices:
\begin{equation} \label{eq:oim_sym}
    \langle V,W \rangle^{(\alpha, \beta)}=\alpha \langle V,W \rangle + \beta \tr(V)\tr(W),
\end{equation}
where $(\alpha,\beta) \in \bfst = \{(\alpha,\beta) \in \mathbb{R}^2 \mid \min (\alpha, \alpha+n \beta)>0\}$, and $V,W \in \sym{n}$.

In this study, we focus on $\biparamLEM$ and LCM. 
We first make some notations and then summarize all the necessary Riemannian operators in \cref{tab:riem_operators}.
Given SPD matrices $P, Q \in \spd{n}$ along with tangent vectors $V, W \in T_P\spd{n}$, we introduce the following notations.
Specifically, the Riemannian metric at $P$ is represented as $g_P(\cdot,\cdot)$, while $\rielog_P(\cdot)$ denotes the Riemannian logarithm at $P$. 
$\pt{P}{Q}$ signifies the parallel transport along the geodesic connecting $P$ and $Q$. 
The matrix exponential and logarithmic functions are denoted as $\mexp(\cdot)$ and $\mlog(\cdot)$, respectively.
In addition, $\chol(\cdot)$ denotes the Cholesky decomposition, with $L=\chol{P}$ and $K=\chol{Q}$ representing the Cholesky factors of $P$ and $Q$. 
The differentials of $\mlog$ and $\chol^{-1}$ at $P$ and $L$ are respectively denoted as $\mlog_{*,P}$ and $(\chol)^{-1}_{*,L}$.
$\lfloor \cdot \rfloor$ refers to the strictly lower part of a square matrix, and $\dlog(L)$ denotes a diagonal matrix comprised of the logarithm of the diagonal elements of $L$.

\begin{table*}[t]
    \centering
    \resizebox{0.99\linewidth}{!}{
    \begin{tabular}{cccc}
         \toprule
         Name & $g_P(V,W)$ & $\rielog_P Q$ & $\Gamma_{P \rightarrow Q} (V)$ \\
         \hline
         $\biparamLEM$ & 
         $\langle \mlog_{*,P} (V), \mlog_{*,P} (W) \rangle^{\alphabeta}$ &
         $(\mlog_{*,P})^{-1} \left[ \mlog(Q) - \mlog(P) \right]$  & 
         $(\mlog_{*,Q})^{-1} \circ \mlog_{*,P} (V)$\\ 
         \hline
         LCM & $\sum_{i>j} \tilde{V}_{i j} \tilde{W}_{i j}+\sum_{j=1}^n \tilde{V}_{j j} \tilde{W}_{j j} L_{j j}^{-2}$ & 
         $ (\chol^{-1})_{*, L} \left[ \lfloor K\rfloor-\lfloor L\rfloor+\bbD(L) \dlog (\bbD(L)^{-1} \bbD(K)) \right]$ & 
         $(\chol^{-1})_{*, K} \left[\lfloor \tilde{V} \rfloor+\bbD(K) \bbD(L)^{-1} \bbD(\tilde{V}) \right]$\\
         \bottomrule
    \end{tabular}  
    }
    \caption{Riemannian operators of $\biparamLEM$ and LCM on SPD manifolds.}
    \label{tab:riem_operators}
    \vspace{-3mm}
\end{table*}

Following the terminology in \cite{chen2023adaptive}, we define the pullback metrics from Euclidean spaces by diffeomorphisms as the  Pullback Euclidean Metrics (PEMs).
Chen~\etal~\cite{chen2023adaptive} demonstrate that both LEM and LCM are PEMs.
We recall an excerpt from Theorem 4.2 of \cite{chen2023adaptive}, covering the properties of PEMs on SPD manifolds.

\begin{theorem}[Pullback Euclidean Metrics (PEMs)] \label{thm:g_spd}
    Let $S, S_1,S_2 \in \spd{n}$ and $ V_1, V_2 \in T_S\spd{n}$, $\phi:\spd{n} \rightarrow \sym{n}$ is a diffeomorphism.
    We define the following operations,
    \begin{align}
        \label{eq:phi_mul} S_1 \phiMul S_2 &= \phiinv( \phi(S_1)+\phi(S_2)),\\
        \label{eq:phi_g} \gphi_S(V_1,V_2) &= \langle \phi_{*,S}(V_1),\phi_{*,S}(V_2) \rangle,
    \end{align}
    where $\phi_{*,S}: T_S\spd{n} \rightarrow T_{\phi(S)}\sym{n}$ is the differential map of $\phi$ at $S$, and $\langle \cdot,\cdot \rangle$ is the standard Frobenius inner product. 
    Then, we have the following conclusions:
    $\{\spd{n}, \phiMul \}$ is an Abelian Lie group, $\{\spd{n}, \gphi \}$ is a Riemannian manifold, and $\gphi$ is a bi-invariant metric, called Pullback Euclidean Metric (\textbf{PEM}).
    The associated geodesic distance is
    \begin{equation} \label{eq:dist_phi_spd}
        \dphi (S_1, S_2 ) = \| \phi(S_1) - \phi(S_2) \|_{\rmF},
    \end{equation}
    where $\|\cdot\|_{\rmF}$ is the norm induced by $\langle \cdot,\cdot \rangle$.
    The Riemannian operators are as follows
    \begin{align}
        \label{eq:gene_rie_exp_spd}
        \rieexp_{S_1} V &= \phiinv(\phi(S_1)+\diffphi{S_1}V),\\
        \label{eq:gene_rie_log_spd}
        \rielog_{S_1}S_2 &= \diffphiinv{\phi(S_1)}(\phi(S_2)-\phi(S_1)),\\
        \label{eq:gene_pt_spd} \pt{S_1}{S_2}(V) &= \diffphiinv{\phi(S_2)} \circ \diffphi{S_1}(V),
    \end{align}
    where $V \in T_{S_1}\spd{n}$ is a tangent vector, $\rieexp_{S_1}$ is the Riemannian exponential at $S_1$, and $\phiinv_{*}$ are the differential maps $\phiinv$.
\end{theorem}
\section{SPD MLRs on SPD manifolds}
\label{sec:rmlr_pems}
This section first reformulates the Euclidean MLR.
Then, we deal with SPD MLR under arbitrary PEM on SPD manifolds.

\subsection{Reformulation of Euclidean MLR} 
\label{subsec:reform_emlr}
The Euclidean MLR was first reformulated in \cite{lebanon2004hyperplane} from the perspective of distances to margin hyperplanes. Hyperbolic MLR was designed based on this reformulation~\citep{ganea2018hyperbolic}. 
In \cite{nguyen2023building}, the authors further proposed three gyro SPD MLRs based on the gyro-structures induced by AIM, LEM, and LCM.
We now briefly review the reformulation of Euclidean MLR.

Given $C$ classes, MLR in $\bbR{n}$ computes the following softmax probabilities:
\begin{equation} \label{eq:EMLR_reform_start}
    \small
    \forall k \in\{1, \ldots, C\}, p(y=k \mid x) \propto \exp \left(\left(\left\langle a_k, x\right\rangle-b_k\right)\right),
\end{equation}
where $b_k \in \mathbb{R}$, and $x, a_k \in \mathbb{R}^n$.
As shown in \cite[Sec. 5]{lebanon2004hyperplane} and \cite[Sec. 3.1]{ganea2018hyperbolic}, \cref{eq:EMLR_reform_start} can be reformulated as
\begin{equation}
    \label{eq:EMLR_reform_end}
    \begin{aligned} 
    & p(y=k \mid x) \propto \\
    & \exp (\operatorname{sign}(\langle a_k, x-p_k\rangle)\|a_k\| d (x, H_{a_k, p_k})), 
\end{aligned}
\end{equation}
where $\langle a_k,p_k \rangle =b_k$, and $H_{a_k, p_k}$ is referred to a hyperplane, defined as
\begin{equation} \label{eq:euc_hyperplane}
    H_{a_k, p_k}=\{x \in \mathbb{R}^n:\langle a_k, x - p_k\rangle=0\}, 
\end{equation}

Recalling \cref{tb:reinter_riem_operators}, $\rielog_p x$ is the natural generalization of the directional vector $\vec{px}=x-p$ starting at $p$ and ending at $x$, while the Riemannian metric at $p$ corresponds to the inner product.
Therefore, the MLR in \cref{eq:EMLR_reform_end} and hyperplane in \cref{eq:euc_hyperplane} can be readily generalized into the SPD manifold $\{\spd{n}, g\}$.

\begin{definition}[SPD hyperplanes] \label{def:spd_hyperplanes}
    Given $P \in \spd{n}, A \in T_P\spd{n} \backslash \{ \bzero \}$, we define the SPD hyperplane as
    \begin{equation} \label{eq:spd_hyperplane}
        \begin{aligned}
            \tilde{H}_{A, P} 
            &=\{S \in \spd{n}: g_{P}( \rielog_P S, A) \\
            &= \langle \rielog_P S, A \rangle_P=0\},
        \end{aligned}
    \end{equation}
    where $P$ and $A$ are referred to as shift and normal matrices, respectively.
    \label{def:spd_hyperplane}
\end{definition}

\begin{definition}[SPD MLR] 
\label{def:spd_mlr}
SPD MLR is defined as
\begin{equation} \label{eq:spd_mlr}
\small
\begin{aligned}
    &p(y=k \mid S) \\
    &\propto \exp (\operatorname{sign}(\langle A_k, \rielog_{P_k}(S) \rangle_{P_k})\|A_k\|_{P_k} d (S, \tilde{H}_{A_k, P_k})),
\end{aligned}
\end{equation}
where $P_k \in \spd{n}$, $A_k \in T_{P_k}\spd{n} \backslash \{\bzero\}$, 
$\langle \cdot, \cdot \rangle_{P_k}=g_{P_k}$, and $\| \cdot \|_{P_k}$ is the norm on $T_{P_k}\spd{n}$ induced by $g$ at $P_k$, and $\tilde{H}_{A_k, P_k}$ is a margin hyperplane in $\spd{n}$ as defined in \cref{eq:spd_hyperplane}.
$d (S, \tilde{H}_{A_k, P_k})$ denotes the margin distance between $S$ and SPD hyperplane $\tilde{H}_{A_k, P_k}$, which is formulated as:
\begin{equation} \label{eq:spd_pem_dist_hyperplane}
    d (S, \tilde{H}_{A_k, P_k})) =\inf _{Q \in \tilde{H}_{A_k, P_k}} d(S, Q),
\end{equation}
where $d(S, Q)$ is the geodesic distance induced by $g$.
\end{definition}

In geometry, the hyperplane in \cref{eq:euc_hyperplane} is actually a regular submanifold of the trivial manifold $\bbR{n}$.
As for our definition of SPD hyperplanes, we have a similar result.

\begin{proposition} [Submanifolds] \label{prop:hyperplanes_as_submanifolds}
    The SPD hyperplane (as defined in \cref{eq:spd_hyperplane}) under any geometrically complete Riemannian metric $g$ is a regular submanifold of SPD manifolds.
\end{proposition}
\begin{proof}
    The proof is presented in \cref{app:subsec:proof_hyperplanes_as_submanifolds}.
\end{proof}

\cref{prop:hyperplanes_as_submanifolds} rationalizes our \cref{def:spd_hyperplanes}, as both the SPD hyperplane and Euclidean hyperplane are submanifolds.
Nevertheless, we still follow the nomenclature of \cite{ganea2018hyperbolic,lebanon2004hyperplane} and call $\tilde{H}_{A, P}$ SPD hyperplane.

\begin{remark} [Difference with the gyro SPD MLR]
Although the gyro SPD MLR introduced in \cite{nguyen2023building} and our method both extend the Euclidean MLR into SPD manifolds, there exist two main differences:
\begin{enumerate}
    \item 
    \textbf{The mathematical techniques employed are different.}
    \cite{nguyen2023building} adopted gyro structures to reformulate \cref{eq:EMLR_reform_end,eq:euc_hyperplane}.
    However, their gyro structures are induced by the Riemannian metrics. Also, the gyro inner product and gyro norm \citep[Def. 2.15]{nguyen2023building} are defined by the inner product and norm in the tangent space at the identity matrix, \ie $T_I\spd{n}$.
    In contrast, our approach directly applies Riemannian geometry to reformulate Euclidean MLR.
    \item 
    \textbf{The margin distance in \cref{eq:spd_pem_dist_hyperplane} are calculated differently.}
    The margin distance in gyro SPD MLR shares the same expression as our \cref{eq:spd_pem_dist_hyperplane}, except that the distance in the right-hand side is gyro distance, which is defined by the distance on $T_I\spd{n}$.
    To bypass the optimization problem in \cref{eq:spd_pem_dist_hyperplane}, Xuan Son Nguyen and Shuo Yang \cite{nguyen2023building} introduced the \textit{pseudo-gyrodistance}.
    In contrast, we directly use the geodesic distance, which is the most natural descriptor for characterizing the distance on manifolds.
\end{enumerate}
\end{remark}

\subsection{SPD MLRs under PEMs}
\label{subsec:general_spd_mlr}
Recalling that for our SPD MLR in \cref{def:spd_mlr}, under most Riemannian metrics on SPD manifolds, all the involved operators in \cref{eq:spd_mlr} have close form expressions, except the margin distance in \cref{eq:spd_pem_dist_hyperplane}.
Therefore, the only difficulty lies in the calculation of the margin distance. This subsection follows the notations in \cref{thm:g_spd} and proposes a general expression for SPD MLRs under PEMs.

We chose PEMs as our starting metrics mainly because of its extensive inclusion and easy computation. Several Riemannian metrics, including LEM, LCM, and their variants \cite{thanwerdas2023n,thanwerdas2022geometry}, all belong to PEMs. Besides, due to the fast and simple calculation of PEMs, the margin distance under a PEM has a closed-form expression, while other metrics like AIM would be complicated to obtain the distances to hyperplanes.

We start by calculating the margin distance in \cref{eq:spd_pem_dist_hyperplane} under a given PEM.
\begin{lemma} \label{lem:dist_to_hyperplane_pems}
    Given a PEM $g$, the margin distance defined in \cref{eq:spd_pem_dist_hyperplane} has a closed-form solution:
    \begin{align} 
        d (S, \tilde{H}_{A_k, P_k})) 
        &= d(\phi(S),H_{\phi_{*,P_k}(A_k),\phi(P_k)}), \\
        \label{eq:dist_s_to_hyperplane}
        &=\frac{|\langle \phi(S)-\phi(P_k),\phi_{*,P_k}(A_k) \rangle|}{\|A_k\|_{P_k}},
    \end{align}
    where $|\cdot|$ is the absolute value.
\end{lemma}
\begin{proof}
    The proof is presented in \cref{app:subsec:proof_dist_to_hyperplane_pems}.
\end{proof}
Putting \cref{eq:dist_s_to_hyperplane} into \cref{eq:spd_mlr}, we obtain our SPD MLR under a given PEM:
\begin{align} 
    p(y=k \mid S) 
    &\propto \exp( \langle A_k, \rielog_{P_k}(S) \rangle_{P_k}),\\
    \label{eq:spd_dist_s_to_h}
    &= \exp (\langle \phi(S)-\phi(P_k),\phi_{*,P_k}(A_k) \rangle),
\end{align}
where $S, P_k \in \spd{n}$ and $A_k \in T_{P_k}\spd{n} \backslash \{\bzero\}$.
When $P_k$ is fixed, $A_k \in T_{P_k}\spd{n}$ indeed lies in a Euclidean space.
However, $P_k$ would vary during training, making $A_k$ non-Euclidean.
To remedy this issue, we propose two solutions.
The first one is the parallel transportation from a fixed tangent space, writing $A_k = \pt{Q}{P_k}(\tilde{A}_k)$ with $\tilde{A}_k \in T_{Q}\spd{n}$ as a Euclidean parameter.
This is the solution also adopted by HNNs \citep{ganea2018hyperbolic}, where the tangent point is the zero vector.
Alternatively, one can also rely on the differential of a Lie group translation, which is widely used in differential manifolds \cite[§ 20]{loring2011introduction}.
Since the Lie groups associated with PEMs are abelian, we only consider the left translation.
We have the following two lemmas to show the relation between the parallel transport and the differential of left translation.

\begin{lemma} \label{lem:equ_pt_and_lt}
    Given a PEM, any parallel transportation is equivalent to the differential map of a left translation and vice versa.
\end{lemma}
\begin{proof}
    The proof is presented in \cref{app:subsec:proof_equ_pt_and_lt}.
\end{proof}

\begin{lemma} \label{lem:pt_anchor_invariance}
    Given two fixed SPD matrices $Q_1,Q_2 \in \spd{n}$, we have the following equivalence for parallel transportations under a PEM,
    \begin{equation}
        \begin{aligned}
            &\forall \tilde{A}_{1,k} \in T_{Q_1}\spd{n}, \exists ! \tilde{A}_{2,k} \in T_{Q_2}\spd{n}, \\
            &s.t. \pt{Q_1}{P_k}(\tilde{A}_{1,k}) = \pt{Q_2}{P_k}(\tilde{A}_{2,k}).
        \end{aligned}
    \end{equation}
\end{lemma}
\begin{proof}
    The proof is presented in \cref{app:subsec:proof_pt_anchor_invariance}.
\end{proof}

\cref{lem:equ_pt_and_lt} indicates that under PEMs, the above two solutions are equivalent, while \cref{lem:pt_anchor_invariance} implies that anchor points can be arbitrarily chosen.
Therefore, without loss of generality, we generate $A_k$ from the tangent space at the identity matrix $I$ by parallel transportation, \ie $A_k = \pt{I}{P_k}(\tilde{A}_k)$ with $\tilde{A}_k \in T_I\spd{n} \cong \sym{n}$.
Together with \cref{eq:gene_pt_spd}, \cref{eq:spd_dist_s_to_h} can be further simplified.

\begin{theorem}[SPD MLR under a PEM] \label{thm:general_mlr_pems}
    Under any PEM, SPD MLR and SPD hyperplane is
    {\small
    \begin{align}
       \label{eq:spd_dist_s_to_h_final}
        p(y&=k \mid S) 
        \propto \exp (\langle \phi(S)-\phi(P_k),\phi_{*,I}(\tilde{A}_k) \rangle),\\
        \label{eq:spd_hyperplane_final}
        \tilde{H}_{\tilde{A}_{k}, P_k} 
        &=\{S \in \spd{n}: \langle \phi(S)-\phi(P_k),\phi_{*,I}(\tilde{A}_k) \rangle=0\},
    \end{align}}
    where $\tilde{A}_{k} \in T_I\spd{n}/\{0\} \cong \sym{n}/\{0\}$ is a symmetric matrix, and $P_k \in \spd{n}$ is an SPD matrix.
\end{theorem}
\begin{proof}
    The proof is presented in \cref{app:subsec:proof_general_mlr_pems}.
\end{proof}

\section{SPD MLRs under deformed LEM and LCM} 
\label{sec:rmlr_deforms}
In this section, 
we first review the deformed LEM and LCM, and then we showcase our SPD MLR in \cref{thm:general_mlr_pems} under these deformed metrics.

Inspired by the deforming utility of the matrix power function \citep{thanwerdas2019affine,thanwerdas2022geometry}, Chen~\etal~\cite{chen2024liebn} define $\triparamLEM$ and $\paramLCM$ as the pullback metric of $\biparamLEM$ and LCM by matrix power function $(\cdot)^\theta$ and scaled by $\frac{1}{\theta^2}$ ($\theta \neq 0$).
As shown in \cite[Props. 5.1]{chen2024liebn}, $\triparamLEM$ is equal to $\biparamLEM$ and $\paramLCM$ interpolates between the standard LCM ($\theta=1$) and an LEM-like metric ($\theta \rightarrow 0$).

Besides, both $\biparamLEM$ and $\paramLCM$ are PEMs \cite{chen2024liebn}.
Therefore, the SPD MLRs under these two families of metrics can be directly obtained by \cref{thm:general_mlr_pems}.

\begin{corollary} [SPD MLRs under the deformed LEM and LCM] 
\label{cor:spd_mlr_param_lem_lcm}
    The SPD MLRs under $\biparamLEM$ is
    \begin{equation}
        \label{eq:spd_mlr_param_lem}
        \begin{aligned}
         & p(y=k \mid S) \propto\\
         & \exp \left [ \langle \mlog(S)-\mlog(P_k), \tilde{A}_{k} \rangle^{\alphabeta} \right ],
        \end{aligned}
    \end{equation}
    where $\tilde{A}_k \in T_I\spd{n} \cong \sym{n} $ and $P_k \in \spd{n}$. The SPD MLRs under $\paramLCM$ is
    \begin{equation}
    \label{eq:spd_mlr_param_lcm}
        p(y=k \mid S) \propto
         \exp \left[ \frac{1}{\theta} \langle X, Y\rangle \right],
    \end{equation}
    with $X$ and $Y$ defined as
    \begin{align}
        X &= \lfloor \tilde{K} \rfloor - \lfloor \tilde{L}_k \rfloor + \left[\dlog(\bbD(\tilde{K}))-\dlog(\bbD(\tilde{L}_k))\right],\\
        Y &= \lfloor \tilde{A}_k \rfloor + \frac{1}{2}\bbD(\tilde{A}_k),
    \end{align}    
    where $\tilde{K}= \chol(S^\theta)$, $\tilde{L}_k = \chol(P_k^\theta)$, and $\bbD(\tilde{A}_k)$ denotes a diagonal matrix with diagonal elements of $\tilde{A}_k$.
\end{corollary} 
\begin{proof}
    The proof is presented in \cref{app:subsec:proof_spd_mlr_param_lem_lcm}.
\end{proof}

\begin{figure}[htbp]
\centering
\includegraphics[width=\columnwidth]{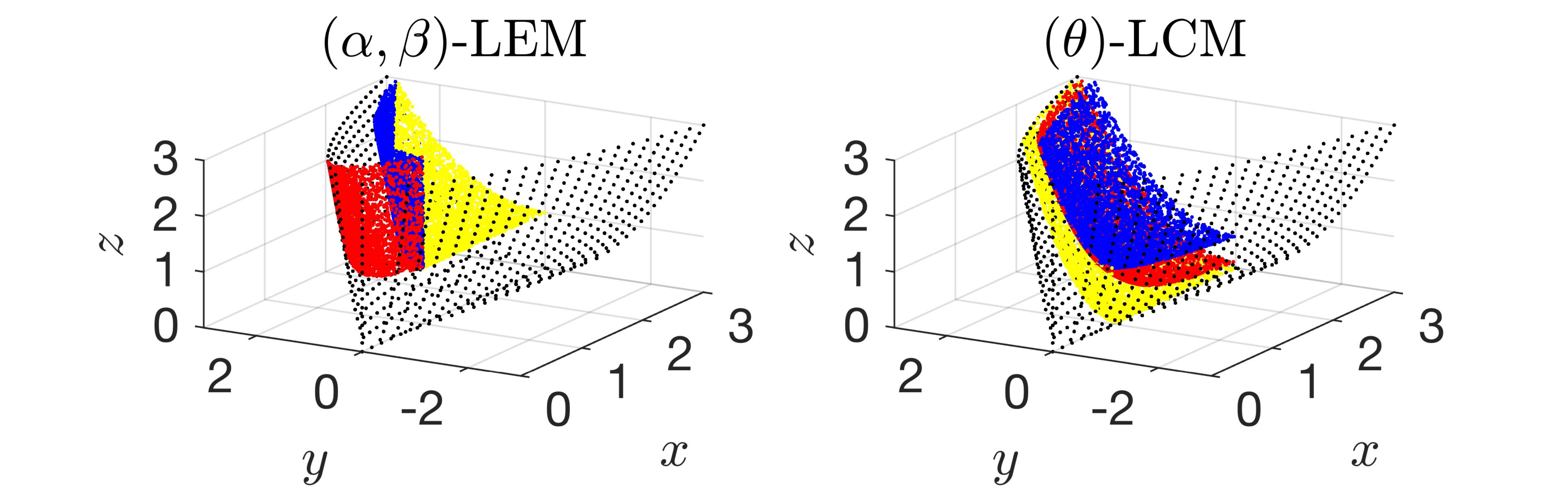}
\caption{Conceptual illustration of SPD hyperplanes induced by $\biparamLEM$ and $\paramLCM$.
In each subfigure, the black dots are symmetric positive semi-definite (SPSD) matrices, denoting the boundary of $\spd{2}$, while the blue, red, and yellow dots denote three SPD hyperplanes.
}
\label{fig:hyperplanes}
\end{figure}

$\spd{2}$ can be visualized as open cone in $\bbR{3}$ by the condition that $\forall P = \left(\begin{array}{ll}
x & y \\
y & z
\end{array}\right)  \in \sym{2}$ is positive definite iff $x,z>0 \land xz >y^2$. 
\cref{fig:hyperplanes} illustrates SPD hyperplanes induced by $\biparamLEM$ and $\paramLCM$.

\begin{remark}
    Our paper incorporates the results w.r.t. LEM and LCM presented in \cite{nguyen2023building}.
    For $\biparamLEM$, when $\alphabeta=(1,0)$, $\biparamLEM$ becomes the standard LEM.
    Our margin distance to the hyperplane in \cref{lem:dist_to_hyperplane_pems} becomes the pseudo-gyrodistance under LEM \citep[Thm. 2.23]{nguyen2023building}.
    For $\paramLCM$, when $\theta=1$, $\paramLCM$ becomes the standard LCM.
    Our \cref{lem:dist_to_hyperplane_pems} becomes the pseudo-gyrodistance induced by LCM \citep[Thm. 2.24]{nguyen2023building}.
    However, our framework does not require gyro structures and directly obtains margin distance and SPD MLR based on the Riemannian metric.
\end{remark}

\section{Rethinking the existing LogEig classifier}
Many of the existing SPD neural networks \citep{huang2017riemannian,brooks2019riemannian,nguyen2019neural,wang2021symnet,nguyen2021geomnet,wang2022dreamnet,chen2023riemannian} rely on a Euclidean MLR in the codomain of matrix logarithm, \ie matrix logarithm followed by an FC layer and a softmax layer. For simplicity, we call this classifier as LogEig MLR. The existing explanation of LogEig MLR is approximating manifolds by tangent space.
However, our framework can offer a novel intrinsic explanation for this widely used MLR.

When $\alphabeta=(1,0)$ for $\biparamLEM$,  the SPD MLR in \cref{eq:spd_mlr_param_lem} is very similar to the LogEig MLR.
However, due to the nonlinearity of $\mlog(\cdot)$ and the non-Euclideanness of SPD parameter $P_k$, SPD MLR cannot be hastily viewed as equivalent to LogEig MLR.
Nevertheless, under special circumstances, \cref{eq:spd_mlr_param_lem} is indeed equivalent to a LogEig MLR.

\begin{proposition} \label{prop:equivalence_lem_mlr}
     Endowing SPD manifolds with the standard LEM, optimizing SPD parameter $P_k$ in \cref{eq:spd_mlr_param_lem} by LEM-based RSGD and Euclidean parameter $A_k$ by Euclidean SGD, the LEM-based SPD MLR is equivalent to a LogEig MLR with parameters in FC layer optimized by Euclidean SGD.
\end{proposition}
\begin{proof}
    The proof is presented in \cref{app:subsec:proof_equivalence_lem_mlr}.
\end{proof}

\cref{prop:equivalence_lem_mlr} implies that optimized by LEM-based RSGD, the LEM-based SPD MLR is equivalent to the Euclidean MLR in the codomain of matrix logarithm.
Nevertheless, a substantial body of prior works underscores the theoretical and empirical superiority of the AIM-based optimization over its LEM-based counterpart \citep{sra2015conic,han2021riemannian}.
Therefore, we adopt the AIM-based optimizer in this paper to update the involved SPD parameters.
\section{Experiments}
\label{sec:experiments}
In this section, we implement the proposed two families of SPD MLRs to SPD neural networks.
Note that our SPD MLRs are architecture-agnostic and can be applied to any existing SPD neural network.
This paper focuses on two network architectures, SPDNet \citep{huang2017riemannian} and TSMNet+SPDDSMBN \citep{kobler2022spd}.
SPDNet is the most classic SPD neural network.
Following previous works \citep{huang2017riemannian,brooks2019riemannian}, we evaluate our SPD MLRs under this architecture for radar recognition on the Radar dataset \citep{brooks2019riemannian} and human action recognition on the HDM05 \citep{muller2007documentation}.
TSMNet+SPDDSMBN is the SOTA Riemannian approach to EEG classification, which is the improved version of SPDNetBN \citep{brooks2019riemannian} for transfer learning on EEG tasks.
We evaluate our SPD MLRs under this baseline for EEG classification on the Hinss2021 dataset \citep{hinss_eegdata_2021}.

\subsection{Baseline models} 
SPDNet \citep{huang2017riemannian} mimics the conventional densely connected feedforward network, consisting of three basic building blocks
\begin{align}
    \text{BiMap: }& S^{k} = W^k S^{k-1} W^k,\\
    \text{ReEig: }&S^{k}=U^{k-1} \max (\Sigma^{k-1}, \epsilon I_{n}) U^{k-1 \top},\\
    \text{LogEig: }&S^{k}=\mln(S^{k-1}),
\end{align}
where $S^{k-1}=U^{k-1} \Sigma^{k-1} U^{k-1 \top}$ is the eigendecomposition, and $W^k$ is semi-orthogonal.
The BiMap (Bilinear Mapping) is the counterpart of linear mapping in Euclidean networks.
The ReEig (Eigenvalue Rectification) mimics the ReLu-like nonlinear activation functions by eigen-rectification.
The LogEig layer projects SPD-valued data into the tangent space for further classification.

The architecture of TSMNet+SPDDSMBN \citep{kobler2022spd} can be explained as 
\begin{equation}
\begin{aligned}
    f_{TC} \rightarrow f_{SC} \rightarrow f_{BiMap} \rightarrow f_{ReEig}\\ \rightarrow f_{SPDDSMBN} \rightarrow f_{LogEig},
\end{aligned}
\end{equation}
where $f_{TC}$ and $f_{SC}$ denote temporal and spatial convolution, and
$f_{SPDDSMBN}$ denotes SPD domain-specific momentum batch normalization, which is an SPD batch normalization layer for domain adaptation. 
For simplicity, we abbreviate TSMNet+SPDDSMBN as SPDDSMBN.

\subsection{Datasets}

\textbf{Datasets and preprocessing: }
\textbf{Radar} dataset \citep{brooks2019riemannian} contains 3,000 synthetic radar signals.
Following the protocol in \cite{brooks2019riemannian}, each signal is split into windows of length 20, resulting in 3,000 covariance matrices of the size $20 \times 20$ equally distributed in 3 classes.
\textbf{HDM05} dataset \citep{muller2007documentation} consists of 2,273 skeleton-based motion capture sequences executed by different actors.
Each frame can be represented as a $93 \times 93$ covariance matrix. 
In line with \cite{brooks2019riemannian}, we remove some under-represented clips and trim the dataset down to 2086 instances scattered throughout 117 classes.
\textbf{Hinss2021} dataset \citep{hinss_eegdata_2021} is a recently released competition dataset containing EEG signals for mental workload estimation.
The dataset is employed for two tasks, namely inter-session and inter-subject classification, which are treated as domain adaptation problems.
Recently, geometry-aware methods \citep{yair2019parallel,kobler2022spd} have demonstrated promising performance in EEG classification, due to the invariance to linear mixing of latent sources and interpretability of SPD modeling \cite{kobler2022spd}. 
We follow the Python implementation\footnote{https://github.com/rkobler/TSMNet} of \citet{kobler2022spd} for data preprocessing.
In detail, the python package MOABB \citep{jayaram_moabb_2018} and MNE \citep{gramfort_meg_2013} are used to preprocess the datasets.
The applied steps include resampling the EEG signals to 250/256 Hz, applying temporal filters to extract oscillatory EEG activity in the 4 to 36 Hz range, extracting short segments ( $\leq 3$s) associated with a class label, and finally obtaining $40 \times 40 $ SPD covariance matrices.

\subsection{Implementation details}

\textbf{Network settings: }
The original classification in SPDNet and TSMNet is conducted by the LogEig MLR (matrix logarithm+FC+softmax). We substitute their LogEig classifiers with our intrinsic SPD MLRs to ensure a fair comparison.
We use the standard cross-entropy loss as the training objective and optimize the parameters with the Riemannian AMSGrad optimizer \citep{becigneul2018riemannian}.
The network architectures are represented as $[d_0, d_1, \ldots, d_L]$, where the dimension of the parameter in the $i$-th BiMap layer is $d_i \times d_{i-1}$.
For the Radar and HDM05 datasets, we adopt a learning rate of $1e^{-2}$, a batch size of 30, and a maximum training epoch of 200. 
For the Hinss2021 dataset, in line with \cite{kobler2022spd}, we apply a learning rate of $1e^{-3}$ with a weight decay of $1e{-4}$, a batch size of 50, and a training epoch of 50.
For better comparison, we also implement the AIM-based gyro SPD MLR \cite{nguyen2023building} to SPDNet and TSMNet, which is named SPDNet+Gyro-AIM or TSMNet+Gyro-AIM. 
All experiments use an Intel Core i9-7960X CPU with 32GB RAM and an NVIDIA GeForce RTX 2080 Ti GPU.

\begin{table}[htpb]
    \centering
    \resizebox{0.9\linewidth}{!}{
    \begin{tabular}{c|c|cc}
        \toprule
        Backbone & Classifier & [20,16,8] & [20,16,14,12,10,8] \\
        \midrule
        \multirow{6}[6]{*}{SPDNet} & LogEig MLR & 92.88±1.05 & 93.47±0.45 \\
        & Gyro-AIM &  94.53±0.95 & 94.32±0.94 \\
        \cmidrule{2-4}          & (1,0)-LEM & 93.55±1.21 & 94.60±0.70 \\
        & (1,1)-LEM & \textbf{95.64±0.83} & \textbf{95.87±0.58} \\
        \cmidrule{2-4}          & (1)-LCM & 93.49±1.25 & 93.93±0.98 \\
        & (0.5)-LCM & \textbf{94.59±0.82} & \textbf{95.16±0.67} \\
        \bottomrule
    \end{tabular}%
    }
    \caption{Results of SPDNet with different classifiers on the Radar dataset.}
    \label{tab:results_radar}
\end{table}

\begin{table}[htpb]
    \centering
    \resizebox{0.99\linewidth}{!}{
    \begin{tabular}{c|c|ccc}
        \toprule
        Backbone & Classifier & [93,30] & [93,70,30] & [93,70,50,30] \\
        \midrule
        \multirow{5}[4]{*}{SPDNet} & LogEig MLR & 57.42±1.31 & 60.69±0.66 & 60.76±0.80 \\
        & Gyro-AIM & 58.07±0.64  &  60.72±0.62 & 61.14±0.94 \\
        \cmidrule{2-5}
        & (1,0)-LEM &  57.02±0.75 &  61.34±0.62 &  60.78±0.86 \\
        \cmidrule{2-5}
        & (1)-LCM & 62.04±1.05 & 62.11±2.11 & 62.89±2.09 \\
        & (0.5)-LCM & \textbf{65.66±0.73} & \textbf{65.79±0.63} & \textbf{65.71±0.75} \\
        \bottomrule
    \end{tabular}%
    }
    \caption{Results of SPDNet with different classifiers on the HDM05 dataset.
    }
    \label{tab:results_HDM05}
    \vspace{-3mm}
 \end{table}

\textbf{Evaluation methods: }
In line with the previous work \citep{huang2017riemannian,kobler2022spd}, we use accuracy as the scoring metric for the Radar and HDM05 datasets, and balanced accuracy (\ie the average recall across classes) for the Hinss2021 dataset. 
Ten-fold experiments on the Radar and HDM05 datasets are carried out with randomized initialization and split, while on the Hinss2021 dataset, models are fit and evaluated with a randomized leave 5\% of the sessions (inter-session) or subjects (inter-subject) out cross-validation scheme.

\textbf{Hyper-parameters: }
We implement the SPD MLRs induced by both the standard metrics and parameterized metrics ($\biparamLEM$ and $\paramLCM$).
Therefore, in our SPD MLRs, we have one or two hyper-parameters, \ie $\theta$ in $\paramLCM$ and $\alphabeta$ in $\biparamLEM$, 
where $\theta$ controls deformation and $\alphabeta$ are associated with $\orth{n}$-invariance.
Recalling \cref{eq:oim_sym}, $\alpha$ is a scaling factor, while $\beta$ measures the relative significance of traces.
As scaling is less important \cite{thanwerdas2019affine}, we set $\alpha=1$.
We select the value of $\beta$ from the candidate set $\{1,\nicefrac{1}{n},\nicefrac{1}{n^2}, 0, -\nicefrac{1}{n} + \epsilon,-\nicefrac{1}{n^2}\}$, where $n$ is the dimension of input SPD matrices in SPD MLRs\footnote{The purpose of including a small positive constant $\epsilon \in \bbRplus$ is to ensure $\orth{n}$-invariance, \ie $\alphabeta \in \bfst$.}. 
These chosen values for $\beta$ allow for amplifying, neutralizing, or suppressing the trace components, depending on the characteristics of the datasets.
For the deformation factor $\theta$, we roughly select its values around the deformation boundary.
Specifically, for $\paramLCM$, $\theta$ is selected from the set $\{ 0.5,1,1.5\}$.

\subsection{Experimental results}

For each family of SPD MLRs, we report two representative baselines: the standard SPD MLR induced from the standard metric ($\theta=1,\alpha=1,\beta=0$), and the one induced from the parameterized metric with selected hyper-parameters. If the standard SPD MLR is already saturated, we only report the results of the standard ones. In \cref{tab:results_radar,tab:results_HDM05,tab:results_inter_subject_session}, we denote $\biparamLEM$ ($\paramLCM$) as the baseline model endowed with the SPD MLR induced by $\biparamLEM$ ($\paramLCM$).

\textbf{Radar: }
In line with \cite{brooks2019riemannian}, we evaluated our classifiers on the Radar dataset under two network architectures: [20, 16, 8] for the 2-layer configuration and [20, 16, 14, 12, 10, 8] for the 5-layer configuration.
The 10-fold results (mean±std) are presented in \cref{tab:results_radar}.
Generally speaking, our SPD MLRs achieve superior performance against the vanilla LogEig MLR.
Among all SPD MLRs, the ones induced by (1,1)-LEM achieve the best performance on this dataset. 
Although the SPD MLRs induced by standard LEM and LCM are slightly worse than the AIM-based gyro SPD MLR, our SPD MLRs with proper hyper-parameters achieve comparable or even better performance than the AIM-based gyro SPD MLR.
For both $\biparamLEM$ ($\paramLCM$), the associated SPD MLR with proper hyper-parameters $\alphabeta$ ($\theta$) outperforms the standard SPD MLR induced by the standard metrics, demonstrating the effectiveness of our parameterization.

\textbf{HDM05: }
Following \cite{huang2017riemannian}, three architectures are evaluated on this dataset: [93, 30], [93, 70, 30], and [93, 70, 50, 30].
The SPD MLR under the standard LEM is already saturated on this dataset and performs similarly to the vanilla LogEig MLR.
This phenomenon might be attributed to the equivalence of a LogEig MLR with an SPD MLR optimized by LEM, which is detailed in \cref{prop:equivalence_lem_mlr}.
Nevertheless, the SPD MLR based on $\paramLCM$ achieves the best performance under different network architectures, improving the vanilla SPDNet by a large margin.
Particularly, (0.5)-LCM demonstrates a clear advantage over the vanilla LogEig MLR.
Besides, the LCM-based SPD MLR consistently performs better than the Gyro-AIM SPD MLR.
More interestingly, the power deformation not only improves the absolute accuracy of the LCM-based SPD MLR but also reduces the standard deviation, indicating the significance of our deformation.
These phenomena demonstrate the advantage of our framework's versatility.



\begin{table}[htbp]
    \centering
    \resizebox{0.9\linewidth}{!}{
    \begin{tabular}{c|c|cc}
    \toprule
    Backbone & Classifier & Inter-session & Inter-subject \\
    \midrule
    \multirow{5}[6]{*}{SPDDSMBN} & LogEig MLR & 53.83±9.77 & 49.68±7.88 \\
          & Gyro-AIM &  53.36±9.92 & 50.65±8.13  \\
    \cmidrule{2-4}          & (1,0)-LEM & 53.16±9.73 & 51.41±7.98 \\
    \cmidrule{2-4}          & (1)-LCM & 55.71±8.57 & 51.60±8.43 \\
          & (1.5)-LCM & \textbf{56.43±8.79} & \textbf{51.65±5.90} \\
    \bottomrule
    \end{tabular}
    }
    \caption{Results of SPDDSMBN with different classifiers on the Hinss2021 dataset under inter-subject and inter-session scenarios.
    The presented results are the ones of balanced accuracy under the leaving 5\% out cross-validation scenario.
    }
    \label{tab:results_inter_subject_session}%
\end{table}%

\textbf{Hinss2021: }
Following \cite{kobler2022spd}, we adopt the architecture of [40,20].
The results (mean±std) of leaving 5\% out cross-validation are presented in \cref{tab:results_inter_subject_session}.
Once again, our intrinsic classifiers demonstrate improved performance compared to the baseline in the inter-session and inter-subject scenarios.
The SPD MLRs based on $\paramLCM$ achieve the best performance (increase 2.6\% for inter-session and 1.97\% for inter-subject), indicating that this metric can faithfully capture the geometry of data in the Hinss2021 dataset.
Besides, the SPD MLR based on powered-deformed LCM shows the least standard deviation compared with other classifiers, demonstrating the significance of the power deformation.
These findings highlight the adaptability and versatility of our framework, as it can effectively leverage different Riemannian metrics based on the intrinsic geometry of the data.

\begin{table}[htbp]
    \centering
    \resizebox{0.99\linewidth}{!}{
    \begin{tabular}{ccccc}
    \toprule
    \multirow{2}[2]{*}{Methods} & \multirow{2}[2]{*}{Radar} & \multirow{2}[2]{*}{HDM05} & \multicolumn{2}{c}{Hinss2021} \\
          &       &       & inter-session & inter-subject \\
    \midrule
    Baseline & 1.36  & 1.95  & 0.18  & 8.31 \\
    \midrule
    MLR-Gyro-AIM & 1.75  & 31.64 & 0.38  & 13.3 \\
    MLR-LEM & 1.5   & 4.7   & 0.24  & 10.13\\
    MLR-LCM & \textbf{1.35}  & \textbf{3.29}  & \textbf{0.18}  & \textbf{8.35} \\
    \bottomrule
    \end{tabular}}
    \caption{Comparison of training efficiency (s/epoch) of SPDNet (SPDDSMBN) under different classifiers. The most efficient MLR is highlighted in \textbf{bold}.}
    \label{tab:training_efficiency}
    \vspace{-3mm}
\end{table}

\subsection{Model efficiency}
We adopt the deepest architectures, namely [20, 16, 14, 12, 10, 8] for the Radar dataset, [93, 70, 50, 30] for the HDM05 dataset, and [40, 20] for the Hinss2021 dataset.
For simplicity, we focus on the SPD MLRs induced by standard metrics, \ie LEM and LCM.
We also implement AIM-based gyro SPD MLR.
The average training time (in seconds) per epoch is reported in \cref{tab:training_efficiency}.
In general, when compared to the AIM-based gyro SPD MLR, LEM- and LCM-based SPD MLRs exhibit superior efficiency, especially when dealing with a larger number of classes. 
Due to the computational complexity of AIM, the AIM-based SPD MLR involves more matrix computation, incurring higher computational costs.
In contrast, due to the rapid computation of PEMs, the PEM-based SPD MLR is more computationally efficient. 
This contrast becomes more obvious when dealing with a huge number of classes, as each class requires an SPD parameter, which needs to be processed by Riemannian computations.
For instance, on the HDM05 dataset, which comprises 117 classes, the LEM- and LCM-based SPD MLRs require only one-ninth training time compared to the AIM-based gyro SPD MLR. 

\subsection{Additional discussions on Gyro SPD MLR}
Theoretically speaking, AIM should generally be more powerful than LEM and LCM, as AIM enjoys affine invariance, which is powerful for modeling covariance matrices.
However, as shown in \cref{tab:results_radar,tab:results_HDM05,tab:results_inter_subject_session}, the improvement of Gyro-AIM SPD MLR for SPDNet is not very significant and is outperformed by our deformed SPD MLR.
This could be attributed to two reasons.
Firstly, as we discussed at the end of \cref{subsec:reform_emlr}, Gyro-AIM MLR did not solve the real margin distance. Instead, its margin distance is defined by Gyro distance, which is defined by the tangent space at the identity. 
However, the most natural distance in manifolds is the geodesic distance. 
Therefore, this might undermine the overall performance of Gyro-AIM MLR. 
On the contrary, our deformed MLRs are developed by the true margin distance.
Secondly, the deformation we adopt can interpolate between different metrics, capturing more vibrant geometry and benefiting our SPD MLR.

Besides, our framework in \cref{thm:general_mlr_pems} enjoys better flexibility than Gyro SPD MLR \cite{nguyen2023building}.
Gyro SPD MLR relies on the gyro structures. 
Given a Riemannian metric, one should first verify whether the induced gyro operations \cite[Eqs. (1-2)]{nguyen2023building} conforms with the 10 axioms of the gyro space \cite[Defs. 2.1 - 2.3]{nguyen2022gyro}.
Besides, one should also solve the induced Gyro SPD MLR \cite{nguyen2023building} based on the verified gyro vector space.
The above process is a case-by-case process.
However, for PEMs, SPD MLRs can be readily obtained by our \cref{thm:general_mlr_pems}, such as the ones induced by LEM, LCM, and their deformed metrics.

\section{Conclusion}
\label{sec:conclusion}
This paper provides a framework for building SPD MLR under any PEM. We showcase our framework under the parameterized metrics of LEM and LCM. Our framework also provides an intrinsic explanation for the widely used LogEig classifier. The consistent superior performance in extensive experiments also supports our claims. As a future avenue, our framework can also be applied to other kinds of PEMs.

\noindent\textbf{Limitations and future work.} This paper constructs SPD MLRs under PEMs, including LEM, LCM, and their variants. 
In the future, we will develop SPD MLRs under other metrics, such as AIM. 
Notably, though AIM-based SPD MLR has been developed \cite{nguyen2023building}, the margin distance is pseudo-gyrodistance, which does not solve \cref{eq:spd_pem_dist_hyperplane}.
Besides, several Euclidean backbones involve SPD features to be classified \cite{wang2017g2denet,engin2018deepkspd,gao2019learning}.
However, the SPD features in these backbones are usually of large dimensions, bringing computational burdens for the Riemannian optimization in our SPD MLR.
In the future, we will explore accelerated optimization to apply our MLR to Euclidean backbones.
\section*{Acknowledgements}
This work was partly supported by the MUR PNRR project FAIR (PE00000013) funded by the NextGenerationEU, the EU Horizon project ELIAS (No. 101120237), and a donation from Cisco. The authors also gratefully acknowledge the financial support from the China Scholarship Council (CSC).

{
    \small
    \bibliographystyle{ieeenat_fullname}
    \bibliography{ref}
}
\clearpage
\appendix
\setcounter{page}{1}
\maketitlesupplementary
\section{Notations} \label{app:notations}
For better understanding, we briefly summarize all the notations used in this paper in \cref{tab:sum_notaitons}.

\begin{table*}[htbp]
    \centering
    \begin{tabular}{cc}
        \toprule
        Notation & Explanation  \\
        \midrule
        $\{\calM , g \}$ or abbreviated as $\calM$ & A Riemannian manifold \\
        $T_P\calM$ & The tangent space at $P \in \calM$\\
        $g_p(\cdot ,\cdot)$ or $\langle \cdot, \cdot \rangle_P$ & The Riemannian metric at $P \in \calM$ \\ 
        $\| \cdot \|_P$ & The norm induced by $\langle \cdot, \cdot \rangle_P$ on $T_P\calM$ \\
        $\rielog_P$ & The Riemannian logarithmic map at $P$\\
        $\rieexp_P$ & The Riemannian exponential map at $P$\\
        $\pt{P_1}{P_2}$ & The Riemannian parallel transportation along the geodesic connecting $P_1$ and $P_2$\\
        $H_{a,p}$ & The Euclidean hyperplane\\
        $\tilde{H}_{\tilde{A}, P}$ & The SPD hyperplane\\
        $\odot$ & A Lie group operation \\
        $\{\calM, \odot\}$ & A Lie group\\
        $P^{-1}_{\odot}$ & The group inverse of $P$ under $\odot$ \\
        $L_P$ & The Lie group left translation by $P \in \calM$ \\
        $f_{*,P}$ & The differential map of the smooth map $f$ at $P \in \calM$\\
        $f^*g$ & The pullback metric by $f$ from $g$\\
        $\spd{n}$ & The SPD manifold \\
        $\sym{n}$ & The Euclidean space of symmetric matrices \\
        $\langle \cdot, \cdot \rangle$ & The standard Frobenius inner product\\
        $\| \cdot \|_\rmF$ & The standard Frobenius norm \\
        $\bfst$ & $\bfst = \{(\alpha,\beta) \in \mathbb{R}^2 \mid \min (\alpha, \alpha+n \beta)>0\}$ \\
        $\langle \cdot, \cdot \rangle^{\alphabeta}$ & The $\orth{n}$-invariant Euclidean inner product \\
        $\mlog$ & Matrix logarithm \\
        $\chol$ & Cholesky decomposition\\
        $\dlog(\cdot)$ & The diagonal element-wise logarithm \\
        $\lfloor \cdot \rfloor$ & The strictly lower triangular part of a square matrix \\
        $\bbD(\cdot)$  & A diagonal matrix with diagonal elements from a square matrix\\
        $\Pi_{P}$ & The tangential projection at $P$ mapping a Euclidean gradient into a Riemannian one\\
        $\nabla_{P} f$ & The Euclidean gradient of $f$ w.r.t. $P$\\
        \bottomrule
    \end{tabular}
    \caption{Summary of notations.}
    \label{tab:sum_notaitons}
\end{table*}

\section{Brief review of Riemannian manifolds}
Intuitively, manifolds are locally Euclidean spaces.
Differentials are the generalization of classical derivatives.
For more details on smooth manifolds, please refer to \cite{loring2011introduction,lee2013smooth}.
Riemannian manifolds are the manifolds endowed with Riemannian metrics, which can be intuitively viewed as point-wise inner products.
When manifolds are endowed with Riemannian metrics, various Euclidean operators can find their counterparts in manifolds.
A plethora of discussions can be found in \cite{do1992riemannian}.

\begin{definition}[Riemannian Manifolds] \label{def:riem_manifold}
A Riemannian metric on $\calM$ is a smooth symmetric covariant 2-tensor field on $\calM$, which is positive definite at every point.
A Riemannian manifold is a pair $\{\calM,g\}$, where $\calM$ is a smooth manifold and $g$ is a Riemannian metric.
\end{definition}

The main paper relies on pullback isometry to study SPD manifolds.
This idea is a natural generalization of bijection from set theory.
\begin{definition} [Pullback Metrics] \label{def:pullback_metrics_app}
    Suppose $\calM,\calN$ are smooth manifolds, $g$ is a Riemannian metric on $\calN$, and $f:\calM \rightarrow \calN$ is smooth.
    Then the pullback of a tensor field $g$ by $f$ is defined point-wisely,
    \begin{equation}
        (f^*g)_p(V_1,V_2) = g_{f(p)}(f_{*,p}(V_1),f_{*,p}(V_2)),
    \end{equation}
    where $p$ is an arbitrary point in $\calM$, $f_{*,p}(\cdot)$ is the differential map of $f$ at $p$, and $V_1,V_2$ are tangent vectors in $T_p\calM$.
    If $f^*g$ is positive definite, it is a Riemannian metric on $\calM$, called the pullback metric defined by $f$.
\end{definition}

\begin{definition}[Isometries] \label{def:isometry}
If $\{M, g\}$ and $\{\widetilde{M}, \widetilde{g}\}$ are both Riemannian manifolds, a smooth map $f: M \rightarrow$ $\widetilde{M}$ is called a (Riemannian) isometry if it is a diffeomorphism that satisfies $f^{*} \tilde{g}=g$.
\end{definition}
If two manifolds are isometric, they can be viewed as equivalent.
Riemannian operators in these two manifolds are also closely related. 

A Lie group is a manifold with a smooth group structure.
It is a combination of algebra and geometry. 
\begin{definition}[Lie Groups] \label{def:lie_group}
A manifold is a Lie group, if it forms a group with a group operation $\odot$ such that $m(x,y) \mapsto x \odot y$ and $i(x) \mapsto x_{\odot}^{-1}$ are both smooth, where $x_{\odot}^{-1}$ is the group inverse of $x$.
\end{definition}

At last, we briefly review the Riemannian gradient.
It is a natural generalization of the Euclidean gradient.
\begin{definition}[Riemannian gradient]
    The Riemannian gradient $\tilde{\nabla} f$ of a smooth function $f \in \cinf(\calM)$ is a smooth vector field  over $\calM$, satisfying
    \begin{equation}
        \langle \tilde{\nabla}_p f, V \rangle_p = V(f), \forall p \in \calM, V \in T_p\calM
    \end{equation}
\end{definition}

\section{Proofs for the lemmas, propositions, theorems, and corollaries stated in the paper} \label{app:sec:proof}

\subsection{Proof of \cref{prop:hyperplanes_as_submanifolds}}
\label{app:subsec:proof_hyperplanes_as_submanifolds}

This claim can be proven by either definition \cite[Def. 9.1]{loring2011introduction} or the constant rank level set theorem \cite[Thm. 11.2]{loring2011introduction}.
We focus on the latter.
\begin{proof}
    Consider any $P \in \spd{n}$ and $A \in T_P \spd{n}$. Define the function $f(S)= \langle \rielog_P S, A\rangle_P :\spd{n} \rightarrow \mathbb{R}$. 
    For the SPD hyperplane $\tilde{H}_{A,P}$, we have $\tilde{H}_{A,P}=f^{-1}(0)$. 
    Due to geodesically completeness, $\rielog_P$ is globally defined, and $f$ is therefore well-defined.
    We can rewrite $f$ as a composition, i.e., $f= h \circ \rielog_P$, where $h(\cdot)=\langle \cdot, A \rangle_P$ is a linear map. 

    Since $\rielog_P$ is a diffeomorphism, and $h(\cdot)$ is a linear map, the rank of $f$ is globally constant. So there exists a neighborhood (e.g., the whole SPD manifold) of $f^{-1}(0)$, where the rank of $f$ is constant. 
    According to the constant rank level set theorem \cite[Thm. 11.2]{loring2011introduction}, we can obtain the claim.
\end{proof}

\subsection{Proof of \cref{lem:dist_to_hyperplane_pems}}
\label{app:subsec:proof_dist_to_hyperplane_pems}

\begin{proof}
    By \cref{thm:g_spd}, we have the following,
    \begin{align} 
        \langle \rielog_P Q, A \rangle_P
        & = \langle \phi_{*,P} \diffphiinv{\phi(P)}(\phi(Q)-\phi(P)), \phi_{*,P} A\rangle\\
        & = \langle \phi(Q)-\phi(P)), \phi_{*,P} A \rangle
    \end{align}
    Therefore, the SPD hyperplane $\tilde{H}_{A_k, P_k}$ corresponds to the Euclidean hyperplane $H_{\phi_{*,P_k}(A_k),\phi(P_k)}$, due to the isometry of $\phi$.
    Furthermore, the distances to margin hyperplanes are equivalent to the following,
    \begin{gather}
         \inf_{\phi(Q)} \| \phi(S) - \phi(Q) \|_\rmF \\ 
         \st \langle \phi(Q) - \phi(P_k), \phi_{*,P_k} A_k \rangle = 0.
    \end{gather}
    The problem above is the familiar Euclidean distance from a point to a hyperplane.
    By simple computation, one can obtain the results.
\end{proof}

\subsection{Proof of \cref{lem:equ_pt_and_lt}}
\label{app:subsec:proof_equ_pt_and_lt}

\begin{proof}
    For simplicity, we abbreviate $\phiMul$ and $\gphi$ as $\odot$ and $g$.
    By abuse of notation, we further denote $Q \odot P^{-1}_{\odot}$ as $Q P^{-1}$, where $P^{-1}_{\odot}$ is the inversion of $P$ under $\odot$.
    According to \cref{thm:g_spd}, $\{\spd{n},\odot\}$ is an Abelian group, $g$ is bi-invariant Riemannian metric.
    By \citet[Lem. 6]{lin2019riemannian}, any parallel transportation can be expressed by a differential of left translation,
    \begin{equation}
        \pt{P}{Q} = L_{Q P^{-1} *,P}, \forall P,Q \in \spd{n}.
    \end{equation}
\end{proof}

\subsection{Proof of \cref{lem:pt_anchor_invariance}}
\label{app:subsec:proof_pt_anchor_invariance}

\begin{proof}
    Due to the geodesic completeness of $\spd{n}$, the existence interval of any geodesic is $\bbRscalar$.
    Parallel transportation along geodesic thus exists for all $t \in \bbRscalar$.
    Through Picard's uniqueness in ODE theories, one can obtain the results.
\end{proof}

\subsection{Proof of \cref{thm:general_mlr_pems}}
\label{app:subsec:proof_general_mlr_pems}

\begin{proof}
    \begin{align}
        A_k &= \pt{I}{P_k}(\tilde{A}_k)\\
            &= \diffphiinv{\phi(P_k)} \circ \diffphi{I}(A_k) \label{eq:pt_i_to_pk}
    \end{align}
    One can obtain the results by putting \cref{eq:pt_i_to_pk} into \cref{eq:spd_dist_s_to_h}.

\end{proof}

\subsection{Proof of \cref{cor:spd_mlr_param_lem_lcm}}
\label{app:subsec:proof_spd_mlr_param_lem_lcm}

\begin{proof}
    Denoting the matrix power as $\pow{\theta}: \spd{n} \rightarrow \spd{n}$, then we have:
    \begin{align} 
        \pow{\theta}(I) &= I, \\
        \label{eq:diff_power_diform_map}
        \pow{\theta *, I}(A) &= \theta A, \forall A \in T_I\spd{n}.
    \end{align}
    Next, we begin to prove the case one by one.

    \textbf{$\biparamLEM$:}
    We define the following map
    \begin{equation} \label{eq:pb_map_def_lem}
        \psi^{\lem}=   f \circ \mlog
    \end{equation}
    where $f:\sym{n} \rightarrow \sym{n}$ is the linear isometry between the standard Frobenius inner product and the $\orth{n}$-invariant inner product $\langle \cdot, \cdot \rangle^{\alphabeta}$.
    Then $\psi^{\lem}$ pulls back the standard Euclidean metric on $\sym{n}$ to $\biparamLEM$ on $\spd{n}$.
    Putting \cref{eq:diff_power_diform_map,eq:pb_map_def_lem} into \cref{eq:spd_dist_s_to_h_final}, we have
    \begin{equation}
        \begin{aligned}
            &\exp (\langle \psi^{\lem}(S)-\psi^{\lem}(P),\psi^{\lem}_{*,I}(\tilde{A}_k) \rangle) \\
            &= \exp \left [ \langle f \left(\mlog(S)-\mlog(P_k) \right), f(\tilde{A}_{k}) \rangle \right ] \\
            &= \exp \left [ \langle  \mlog(S)-\mlog(P_k), \tilde{A}_{k} \rangle^{\alphabeta} \right ],
        \end{aligned}
    \end{equation}
    where the last equation comes from the fact that $f=f_{*}$.

    \textbf{$\paramLCM$:}
    We denote
    \begin{equation}  \label{eq:psi_lcm}
        \psi^{\lcm} =  \dlog \circ \chol  \circ \pow{\theta},
    \end{equation}
    then $\psi^{\lcm}$ pulls back the Euclidean metric $\frac{1}{\theta^2}\geuc$ on the Euclidean space $\tril{n}$ of lower triangular matrices to the $\paramLCM$ on $\spd{n}$.
    The differential of Cholesky decomposition is presented in \citet[Prop. 4]{lin2019riemannian}, while the differential of $\dlog$ can be found in \cite{chen2023adaptive}.
    Then, simple computations show that
    \begin{equation} \label{eq:diff_psi_lcm}
        \psi^{\lcm}_{*,I}(A) = \theta \left( \lfloor A \rfloor + \frac{1}{2}\bbD(A) \right), \forall A \in T_I\spd{n}.
    \end{equation}
    Putting \cref{eq:psi_lcm,eq:diff_psi_lcm} into \cref{eq:spd_dist_s_to_h_final}, we can obtain the results.
\end{proof}

\subsection{Proof of \cref{prop:equivalence_lem_mlr}}
\label{app:subsec:proof_equivalence_lem_mlr}

To prove \cref{prop:equivalence_lem_mlr}, we first present two lemmas about the general cases under PEMs.

One can observe that \cref{eq:spd_dist_s_to_h_final} and \cref{eq:spd_hyperplane_final} are very similar to a Euclidean MLR.
However, since $\phi$ is normally non-linear and $P_k$ is an SPD parameter, \cref{eq:spd_dist_s_to_h_final} cannot hastily be identified with a Euclidean MLR.
However, under some special circumstances, SPD MLR can be reduced to the familiar Euclidean MLR.
To show this result, we first present the Riemannian Stochastic Gradient Descent (RSGD) under PEMs.
General RSGD \citep{bonnabel2013stochastic} is formulated as 
\begin{equation} \label{eq:general_rsgd}
    W_{t+1}=\rieexp _{W_t}(-\gamma_t \Pi_{W_t}( \nabla_W f |_{W_t}))
\end{equation}
where $\Pi_{W_t}$ denotes the projection mapping Euclidean gradient $\nabla_W f|_{W_t}$ to Riemannian gradient, and $\gamma_t$ denotes learning rate.
We have already obtained the formula for the Riemannian exponential map as shown in \cref{eq:gene_rie_log_spd}.
We proceed to formulate $\Pi$.
\begin{lemma} \label{lem:rsgd_pems}
    For a smooth function $f: \spd{n} \rightarrow \bbRscalar$ on $\spd{n}$ endowed with any kind of PEMs, the projection map $\Pi_{P}: \sym{n} \rightarrow T_P\spd{n}$ at $P \in \spd{n}$ is
    \begin{equation} \label{eq:proj_pems}
        \Pi_{P} (\nabla_{P} f) = \phi^{-1}_{*,P} (\phi^{-*}_{*,P}) (\nabla_P f),
    \end{equation}
    where $\phi^{-*}_{*,P}$ is the adjoint operator of $\phi^{-1}_{*,P}$, \ie $\langle V_1, \phi^{-1}_{*,P} V_2 \rangle_P = \langle \phi^{-*}_{*,P} V_1, V_2 \rangle_P$, for all $V_i \in T_P\spd{n}$.
\end{lemma}
\begin{proof}
    Given any smooth function $f:\spd{n} \rightarrow \bbRscalar$, denote its Riemannian gradient at $P$ as $\tilde{\nabla}_{P}f \in T_P\spd{n}$.
    Then we have the following,
    \begin{align}
        \langle \tilde{\nabla}_{P}f, V \rangle_P = V(f), \forall V \in T_P\spd{n}.
    \end{align}
    By \cref{eq:phi_g} and canonical chart, we have
    \begin{equation}
        \langle \phi_{*,P}\tilde{\nabla}_{P}f,\phi_{*,P} V\rangle = \langle \nabla_P f, V \rangle, \forall V \in T_P\spd{n} \cong \sym{n},
    \end{equation}
    where $\nabla_P f$ is the Euclidean gradient.
    By the arbitrary of $V$, we have
    \begin{equation}
        \phi_{*,P}^*\phi_{*,P}\tilde{\nabla}_{P}f =  \nabla_P f,
    \end{equation}
    where $\phi_{*,P}^*$ is the adjoint operator of the linear homomorphism $\phi_{*,P}$ w.r.t. $\langle,\rangle$.
\end{proof}

We can describe the special case we mentioned with the above lemma.
\begin{lemma} \label{lem:spdmlr_to_emlr}
    Supposing the differential map $\phi_{*,I}$ is the identity map, and $P_k$ in \cref{eq:spd_dist_s_to_h_final} is optimized by PEM-based RSGD, then \cref{eq:spd_dist_s_to_h_final} can be reduced to a Euclidean MLR in the codomain of $\phi$ updated by Euclidean SGD.
\end{lemma}
\begin{proof}
    Define a Euclidean MLR in the codomain of $\phi$ as
    \begin{equation}
        p(y=k \mid S) 
        \propto \exp (\langle \phi(S)-\bar{P}_k,\bar{A}_k) \rangle), 
    \end{equation}
    where $\bar{P}_k, \bar{A}_k \in \sym{n}$. We call this classifier $\phi$-EMLR.
    
    Define the SPD MLR under the PEM induced by $\phi$ is 
    \begin{equation}
        p(y=k \mid S) 
        \propto \exp (\langle \phi(S)-\phi(P_k), \tilde{A}_k \rangle), 
    \end{equation}
    where $P_k \in \spd{n}, \tilde{A}_k \in \sym{n}.$
    
    Supposing the SPD MLR and $\phi$-EMLR satisfying $\bar{P}_k = \phi(P_k)$.
    Other settings of the network are all the same, indicating the Euclidean gradients satisfying
    \begin{equation}
        \frac{\partial L}{\partial \bar{P}_k} = \frac{\partial L}{\partial \phi(P_k)}.
    \end{equation}
    The updates of $\bar{P}_k$ in the $\phi$-EMLR is 
    \begin{equation}
        \bar{P}'_k = \bar{P}_k - \gamma \frac{\partial L}{\partial \bar{P}_k}.
    \end{equation}
    The updates of ${P}_k$ in the SPD MLR is 
    \begin{align}
        P'_{k} &=\rieexp _{P_{k}}(-\gamma \Pi_{P_{k}}( \nabla_{P_{k}} f))\\
               &= \phi^{-1} (\phi(P_k) - \gamma \phi_{*,P_k}^{-*} \frac{\partial L}{\partial {P}_k})
    \end{align}
    Therefore $\phi(P'_{k})$ satisfies
    \begin{align}
         \phi(P'_{k}) 
         &= \phi(P_k) - \gamma \phi_{*,P_k}^{-*} \frac{\partial L}{\partial {P}_k}\\
         &= \phi(P_k) - \gamma \phi_{*,P_k}^{-*} \phi_{*,P_k}^* \frac{\partial L}{\partial \phi({P}_k)} \label{eq:phi_p_bp}\\ 
         &= \phi(P_k) - \gamma \frac{\partial L}{\partial \phi({P}_k)}\\
         &= \bar{P}'_{k}
    \end{align}
\cref{eq:phi_p_bp} comes from the Euclidean chain rule of differential.
Let $Y=\phi(X)$, then we have
\begin{align}
    \frac{\partial L}{\partial Y}: \diff Y=\frac{\partial L}{\partial Y}: \phi_{*, X} \diff X=\phi_{*,X}^* \frac{\partial L}{\partial Y}: \diff X,
\end{align}
where $ : $ means Frobenius inner product.

The equivalence of $\bar{A}_k$ and $\tilde{A}_k$ is obvious.
By natural induction, the claim can be proven.
\end{proof}

Now, We can directly prove \cref{prop:equivalence_lem_mlr} by \cref{lem:spdmlr_to_emlr}.


\end{document}